\documentclass{article}
\usepackage{spconf,amsmath,graphicx,hyperref}
\usepackage{amssymb, amsthm}

\usepackage{algorithm,algpseudocode}
\usepackage{stfloats}

\usepackage{booktabs}
\usepackage{multirow}
\usepackage{graphicx}
\usepackage{xcolor}

\newsavebox{\combinedFlowBox}

\usepackage{etoolbox}

\apptocmd{\thebibliography}{%
  \fontsize{9}{11}\selectfont
  \setlength{\itemsep}{0pt}%
  \setlength{\parsep}{0pt}%
  \setlength{\parskip}{0pt}%
}{}{%
  \PackageError{bibformat}{Bibliography patch failed}{}%
}

\title{CLIMB-flow: Coupled Linear Inverse posterior
sampling via Multiscale-Based flow}
\name{Zeqiu Yu$^1$, Ruizhi Yuan$^2$, Mathews Jacob$^1$}

\address{$^1$Department of Electrical and Computer Engineering,
University of Virginia, Charlottesville, VA, USA\\
$^2$Department of Biostatistics and Health Data Science, University of Pittsburgh, Pittsburgh, PA, USA}
\begin{document}
%
\maketitle
\begin{abstract}
Diffusion models are now widely used in Bayesian inverse problems in imaging as priors, where latent diffusion models are often used for larger scale problems to keep the computational complexity and model-size manageable. Unfortunately, the auto-encoder based compression result in loss of spatial detail. In addition, the optimization is converted to a non-linear problem. In this paper, we introduce a posterior sampling algorithm customized for the pyramidal/cascaded architecture, which relies on a coarse to fine heirarchical strategy to generate images in the pixel domain. We present CLIMB-Flow which alternates between three steps: an end-point estimation from the current coarse and noisy image, data-consistent update of the clean image, and re-noising it back
to the level the network expects. Together these steps sample the
posterior at that scale using an approximate Gibbs sampling from two conditional distribution. Experiments on
ImageNet, CelebA, AFHQ and fastMRI span inpainting, deblurring,
super-resolution and accelerated MRI, with PSNR gains of
1.37--7.66\,dB over the strongest competing method on CelebA and
pixel-domain reconstruction up to $512\times512$.

\end{abstract}
\section{Introduction}
\label{sec:intro}

Linear inverse problems are central to scientific and medical imaging, including accelerated MRI, and to vision tasks such as inpainting, deblurring, and super-resolution. These problems are often ill-posed, as measurements alone do not uniquely determine the underlying image. Posterior sampling addresses this ambiguity by combining prior information with the measurement likelihood to generate plausible reconstructions conditioned on the observations. Modern generative models provide expressive, data-driven priors for this purpose~\cite{song2021solving,kawar2022denoising,chung2022diffusion,wang2022zero,zhang2025improving,purohit2025consistency,xia2026noise}. Diffusion-based approaches are widely used~\cite{song2021solving,daras2023solving,park2025measurement}, while flow matching and rectified flow models~\cite{lipman2022flow,liu2022flow,albergo2023building} offer promising transport-based alternatives.

Scaling these methods to high-resolution reconstruction remains challenging because inference repeatedly evaluates generative networks at full resolution. A common approach is to reduce computation through autoencoder-based latent representations~\cite{rombach2022high,labs2025flux1kontextflowmatching}, which introduces two difficulties. First, lossy compression can suppress fine spatial details important to scientific and medical imaging~\cite{daras2024survey,song2023solving,lee2025latent,luo2023hfssde,chung2022bjr}. Second, even when the measurement operator $A$ is linear, the effective map $z\mapsto A\operatorname{Dec}(z)$ is nonlinear in the latent variable, requiring differentiation through the decoder for gradient-based measurement updates.

Multiscale generative models offer a complementary route to efficient high-resolution synthesis. Cascaded diffusion~\cite{ho2022cascaded}, wavelet-based diffusion~\cite{phung2023wavelet}, and pyramidal flow matching~\cite{jin2024pyramidal} exploit hierarchical representations to reduce computational demands. PixelFlow~\cite{chen2025pixelflow} implements cascaded flow modeling directly in pixel space, progressing through increasingly fine resolutions and reserving full-resolution processing for the final stage. This design reduces the cost of early inference without requiring a latent-space decoder. Extending it to inverse problems, however, requires a posterior formulation that reconciles stage-dependent image variables with full-resolution measurements while preserving the linear measurement structure.

We introduce \textbf{CLIMB-Flow} (\textbf{C}oupled \textbf{L}inear \textbf{I}nverse posterior sampling via \textbf{M}ultiscale-\textbf{B}ased Flow), which is a posterior sampler for cascaded pixel-domain flow priors. Our coupled stage-wise posterior links clean images and noisy interpolants through scale-adapted linear measurement models, sampled by Gibbs sampling while annealing the scale (Sec.~\ref{sec:method}). The required linear systems are solved by matrix-free conjugate gradients without network backpropagation; only 20 of 80 network evaluations operate at full resolution.

We also consider a CLIMB-Flow (Mode) variant, where the Gaussian image-conditional draw is replaced by its mean. Mode leads PSNR/SSIM on all four CelebA tasks (Table~\ref{tab:combined_results}); the stochastic sampler leads perceptual quality on ImageNet inpainting and super-resolution. Further experiments cover accelerated fastMRI reconstruction and $512\times512$ AFHQ reconstruction.

\begin{figure}[!htbp]
\vspace{-6pt}
  \centering
  \includegraphics[width=\columnwidth]{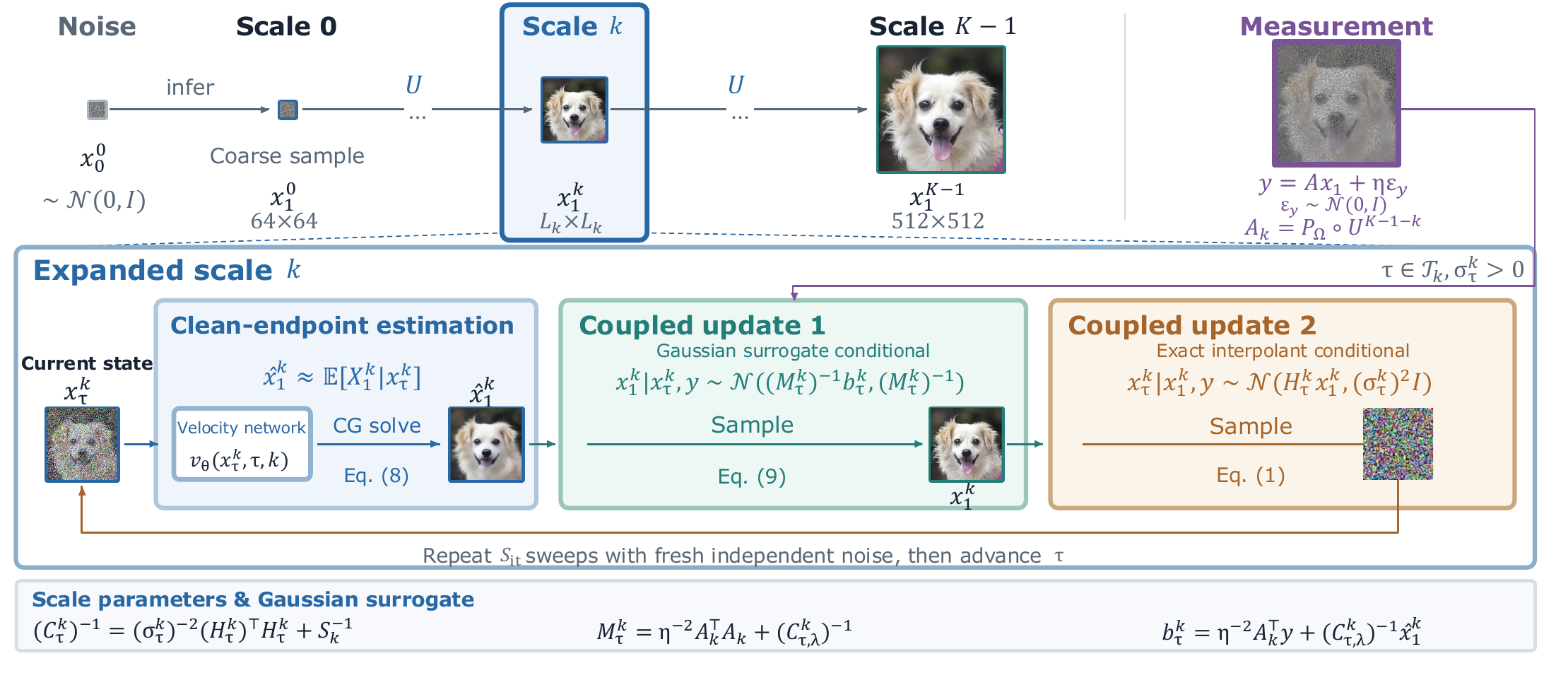}
  \vspace{-21pt}
    \caption{\textbf{Preview of CLIMB-Flow reconstructions.} Inference proceeds from coarse to fine, with upsampling (U) linking successive scales. At scale $k$, a coupled target connects the noisy interpolant $x_\tau^k$ and clean image $x_1^k$: the velocity network supplies prior information through the former, while measurements constrain the latter. Each panel shows one step of the approximate Gibbs sweep, repeated $S_{\mathrm{it}}$ times before advancing to the next scale.}
  \label{fig:Cover}
\vspace{-9pt}
\end{figure}

\section{Background}
\label{sec:background}

PixelFlow~\cite{chen2025pixelflow} splits the transport time $[0,1]$
into $K$ consecutive stages, indexed $k=0,\ldots,K-1$ from coarsest
to finest. Stage $k$ correspond to the time-indices $[s_k,e_k]$, with
$s_0=0$, $e_{K-1}=1$ and $e_k=s_{k+1}$. Within a stage we use the rescaled time
$\tau=(t-s_k)/(e_k-s_k)\in[0,1]$.

To simplify the hierarchical posterior sampler, we reformulate the expression incorporating linear scale operators. Let $D$ and $U$ denote one-step linear downsampling and
upsampling, with $D^j,U^j$ their $j$-fold compositions. At
stage $k$, the clean image is $x_1^k:=D^{K-1-k}x_1$, with law
$p_1^k$, and the noise endpoint is
$x_0^k\sim\mathcal{N}(0,I_{n_k})$, drawn independently. Thus, scale $k$ and works on images of size $n_k = \frac{n}{2^{(K-1-k)}}$ pixels, where $n$ is the size of the original image. We denote 
$G:=U\circ D$ which is a blur function.  Between stages the state is carried forward as $x_1^{k+1}\gets U(x_1^k)$, with a
fresh $x_0^{k+1}$.

Stage $k$ runs from
$x_{\mathrm{start}}^k=s_kGx_1^k+(1-s_k)x_0^k$ to
$x_{\mathrm{end}}^k=e_kx_1^k+(1-e_k)x_0^k$. Interpolating between
them gives
\vspace{-5pt}
\begin{equation}
x_\tau^k = H_\tau^k x_1^k+\sigma_\tau^k x_0^k,
\label{eq:interpolant}
\vspace{-6pt}
\end{equation}
with $H_\tau^k := (1-\tau)s_kG+\tau e_kI$,
$\sigma_\tau^k := (1-\tau)(1-s_k)+\tau(1-e_k)$.
We note that $H_\tau^k$ is symmetric,
invertible for $\tau>0$, and rank-deficient at $\tau=0$, where it
reduces to $s_kG$; $\sigma_\tau^k$ is positive except at $\tau=1$
of the final stage. Subtracting the start from the end gives the
stage displacement given $x_\tau$
\vspace{-6pt}
\begin{equation}
d_\tau^k=x_{\mathrm{end}}^k-x_{\mathrm{start}}^k
=B_kx_1^k-\Delta_kx_0^k,
\label{eq:displacement}
\vspace{-6pt}
\end{equation}
with $B_k=e_kI-s_kG$ and $\Delta_k=e_k-s_k$, predicted by the pretrained network $v(x_\tau^k):=v_\theta(x_\tau^k)$. With information from $y=Ax_1+\eta\varepsilon_y$, $A$ is a known
linear operator, $\varepsilon_y\sim\mathcal{N}(0,I)$ and $\eta>0$ we can recover $x_1$ by matching the measurement at scale $k$ through $A_k=A\circ U^{K-1-k}$, giving the Gaussian likelihood $p(y\mid x_1^k)=\mathcal{N}(y;A_kx_1^k,\eta^2I)$. $A_{K-1}=A$ is exact at full resolution.
\vspace{-12pt}
\newsavebox{\CFImageNetMeasureBox}
\newsavebox{\CFCelebAMeasureBox}
\newlength{\CFSharedTableWidth}

\begin{table*}[t!]
\vspace{-4pt}
\centering
\setlength{\parskip}{0pt}

\begingroup
\scriptsize
\setlength{\tabcolsep}{2pt}
\renewcommand{\arraystretch}{1.10}
\setlength{\aboverulesep}{1pt}
\setlength{\belowrulesep}{1pt}
\setlength{\abovetopsep}{0pt}
\setlength{\belowbottomsep}{0pt}

\newcommand{\CFImageNetRows}{%
\toprule
\multirow{2}{*}{Type} & \multirow{2}{*}{Model}
& \multicolumn{4}{c}{Inpaint (box)}
& \multicolumn{4}{c}{Inpaint (random)}
& \multicolumn{4}{c}{Super resolution 4$\times$} \\
\cmidrule(lr){3-6}
\cmidrule(lr){7-10}
\cmidrule(lr){11-14}
&
& PSNR$\uparrow$ & SSIM$\uparrow$ & LPIPS$\downarrow$ & FID$\downarrow$
& PSNR$\uparrow$ & SSIM$\uparrow$ & LPIPS$\downarrow$ & FID$\downarrow$
& PSNR$\uparrow$ & SSIM$\uparrow$ & LPIPS$\downarrow$ & FID$\downarrow$ \\
\midrule
P;F & Ours (Mode)
& \textbf{22.21} & \textbf{0.834} & \underline{0.188} & \underline{86.80}
& 28.96 & \textbf{0.856} & \underline{0.110} & \underline{35.70}
& \textbf{25.97} & \textbf{0.712} & 0.351 & 97.00 \\

P;F & Ours
& \underline{22.17} & \underline{0.779} & \textbf{0.179} & \textbf{77.80}
& 28.18 & 0.795 & \textbf{0.105} & \textbf{32.40}
& 25.31 & 0.681 & \textbf{0.266} & \textbf{79.49} \\

\midrule
P;D & DAPS \cite{zhang2025improving}
& 21.43 & 0.725 & 0.214 & 109.85
& 28.44 & 0.775 & 0.135 & 54.25
& \underline{25.89} & 0.694 & \underline{0.276} & \underline{83.57} \\

P;D & DPS \cite{chung2022diffusion}
& 18.94 & 0.722 & 0.257 & 126.52
& 23.52 & 0.745 & 0.297 & 87.53
& 21.13 & 0.489 & 0.361 & 106.32 \\

P;D & DDRM \cite{kawar2022denoising}
& 18.63 & 0.733 & 0.254 & 116.37
& -- & -- & -- & --
& 22.62 & 0.521 & 0.324 & 103.85 \\

P;D & DDNM \cite{wang2022zero}
& 21.64 & 0.748 & 0.319 & 103.97
& \underline{31.16} & \underline{0.841} & 0.191 & 63.84
& 23.96 & 0.604 & 0.475 & 98.62 \\

P;D & FPS-SMC \cite{dou2024diffusion}
& 22.16 & 0.726 & 0.208 & 111.58
& 24.52 & 0.701 & 0.316 & 79.12
& 24.82 & \underline{0.703} & 0.313 & 97.51 \\

P;D & DiffPIR \cite{zhu2023denoising}
& -- & -- & -- & --
& -- & -- & -- & --
& 23.18 & -- & 0.371 & 106.32 \\

\midrule
L;D & LatentDAPS \cite{zhang2025improving}
& 17.19 & 0.624 & 0.340 & 145.63
& 27.59 & 0.772 & 0.164 & 61.62
& 25.06 & 0.673 & \underline{0.276} & 84.37 \\

L;D & PSLD \cite{rout2023solving}
& 20.10 & 0.694 & 0.465 & 146.53
& \textbf{31.30} & 0.783 & 0.337 & 83.21
& 25.42 & 0.694 & 0.360 & 97.45 \\

L;D & ReSample \cite{song2023solving}
& 18.29 & 0.631 & 0.262 & 127.84
& 27.50 & 0.756 & 0.143 & 59.87
& 22.61 & 0.576 & 0.370 & 113.42 \\

\bottomrule
}

\newcommand{\CFCelebARows}{%
\toprule
\multirow{2}{*}{Type} & \multirow{2}{*}{Method}
& \multicolumn{3}{c}{Deblurring}
& \multicolumn{3}{c}{Super-resolution}
& \multicolumn{3}{c}{Random inpainting}
& \multicolumn{3}{c}{Box inpainting} \\
\cmidrule(lr){3-5}
\cmidrule(lr){6-8}
\cmidrule(lr){9-11}
\cmidrule(lr){12-14}
&
& PSNR$\uparrow$ & SSIM$\uparrow$ & LPIPS$\downarrow$
& PSNR$\uparrow$ & SSIM$\uparrow$ & LPIPS$\downarrow$
& PSNR$\uparrow$ & SSIM$\uparrow$ & LPIPS$\downarrow$
& PSNR$\uparrow$ & SSIM$\uparrow$ & LPIPS$\downarrow$ \\
\midrule
P;D & DiffPIR \cite{zhu2023denoising}
& 32.77 & 0.912 & 0.060
& 31.52 & 0.895 & 0.033
& 31.74 & 0.917 & 0.025
& -- & -- & -- \\

P;F & OT-ODE \cite{pokle2023training}
& 33.01 & 0.921 & \underline{0.029}
& 31.46 & 0.907 & \underline{0.025}
& 28.68 & 0.871 & 0.051
& 29.40 & 0.920 & 0.038 \\

P;F & D-Flow \cite{ben2024d}
& 31.25 & 0.854 & 0.038
& 30.47 & 0.843 & 0.026
& 33.67 & 0.943 & \textbf{0.015}
& 30.70 & 0.899 & \underline{0.026} \\

P;F & Flow-Priors \cite{zhang2024flow}
& 31.54 & 0.858 & 0.056
& 28.35 & 0.713 & 0.102
& 32.88 & 0.871 & 0.019
& 30.07 & 0.858 & 0.048 \\

P;F & PnP-Flow \cite{martin2025pnp}
& 34.48 & 0.936 & 0.040
& 31.09 & 0.902 & 0.045
& 33.05 & \underline{0.944} & 0.018
& 30.47 & 0.933 & 0.037 \\

P;F & Flower-OT \cite{pourya2025flower}
& 34.98 & \underline{0.947} & \textbf{0.026}
& 32.36 & \underline{0.923} & 0.034
& 33.08 & \underline{0.944} & 0.018
& 31.19 & \underline{0.945} & \textbf{0.022} \\

\midrule
P;F & \textbf{Ours (Mode)}
& \textbf{36.35} & \textbf{0.958} & 0.044
& \textbf{35.61} & \textbf{0.936} & 0.031
& \textbf{37.25} & \textbf{0.966} & 0.026
& \textbf{38.85} & \textbf{0.981} & \underline{0.026} \\

P;F & \textbf{Ours}
& \underline{35.08} & 0.942 & 0.039
& \underline{33.41} & 0.915 & \textbf{0.024}
& \underline{35.23} & 0.932 & \underline{0.016}
& \underline{34.94} & 0.922 & 0.028 \\

\bottomrule
}

\sbox{\CFImageNetMeasureBox}{%
  \begin{tabular}{@{}ll*{12}{c}@{}}
    \CFImageNetRows
  \end{tabular}%
}
\sbox{\CFCelebAMeasureBox}{%
  \begin{tabular}{@{}ll*{12}{c}@{}}
    \CFCelebARows
  \end{tabular}%
}
\setlength{\CFSharedTableWidth}{\wd\CFImageNetMeasureBox}
\ifdim\wd\CFCelebAMeasureBox>\CFSharedTableWidth
  \setlength{\CFSharedTableWidth}{\wd\CFCelebAMeasureBox}
\fi

\noindent
\begin{minipage}[t]{0.495\textwidth}
  \vspace{0pt}
  \centering
  \resizebox{\linewidth}{!}{%
    \begin{tabular*}{\CFSharedTableWidth}{@{\extracolsep{\fill}}ll*{12}{c}@{}}
      \CFImageNetRows
    \end{tabular*}%
  }\par
\end{minipage}\hfill%
\begin{minipage}[t]{0.495\textwidth}
  \vspace{0pt}
  \centering
  \resizebox{\linewidth}{!}{%
    \begin{tabular*}{\CFSharedTableWidth}{@{\extracolsep{\fill}}ll*{12}{c}@{}}
      \CFCelebARows
    \end{tabular*}%
  }\par
\end{minipage}\par
\endgroup

\setlength{\abovecaptionskip}{2pt}
\setlength{\belowcaptionskip}{0pt}
\caption{
\textbf{Quantitative results.}
\textbf{Left:} ImageNet comparisons with pixel- and latent-domain diffusion baselines. Ours (Mode) achieves the highest SSIM on all
three tasks, while Ours achieves the lowest LPIPS/FID on both
inpainting tasks and super-resolution.
\textbf{Right:} CelebA comparisons with pixel-domain
flow solvers and DiffPIR. Ours (Mode) leads PSNR/SSIM on all
four tasks, with PSNR gains of $1.37$--$7.66$\,dB over the
best competing baseline per task; Ours improves LPIPS over
Mode on three tasks.
\textbf{Bold}/\underline{underline}: best/second-best among
all methods in each panel, including both
CLIMB-Flow variants; ties are marked equally.
P/L: pixel/latent domain; D/F: diffusion/flow.
}
\label{tab:combined_results}
\vspace{-6pt}
\end{table*}

\begin{figure*}[b!]
\vspace{-4pt}
\centering
\setlength{\abovecaptionskip}{2pt}
\setlength{\belowcaptionskip}{0pt}
\noindent
\begin{minipage}[t]{0.495\textwidth}
  \centering
  \includegraphics[width=1.2\linewidth,keepaspectratio]{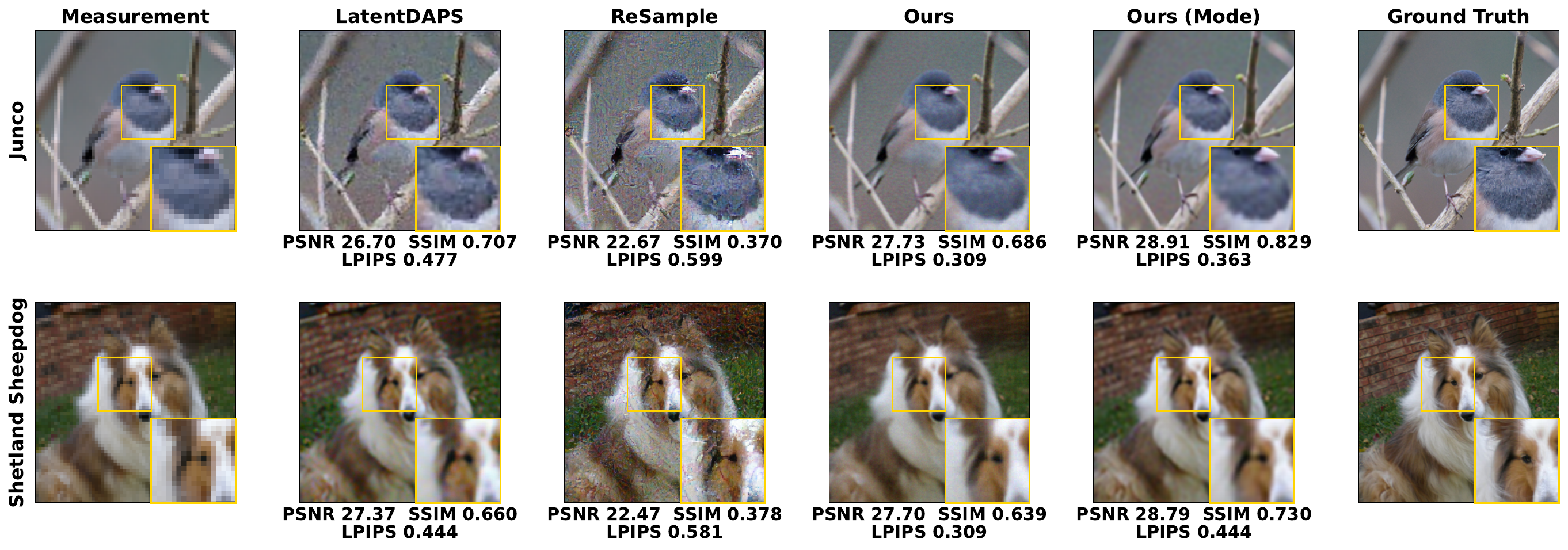}\par
  \small (a)
\end{minipage}\hfill%
\begin{minipage}[t]{0.405\textwidth}
  \centering
  \includegraphics[
    width=\linewidth,
    keepaspectratio,
    trim={0pt 0pt 0pt 0pt},
    clip
  ]{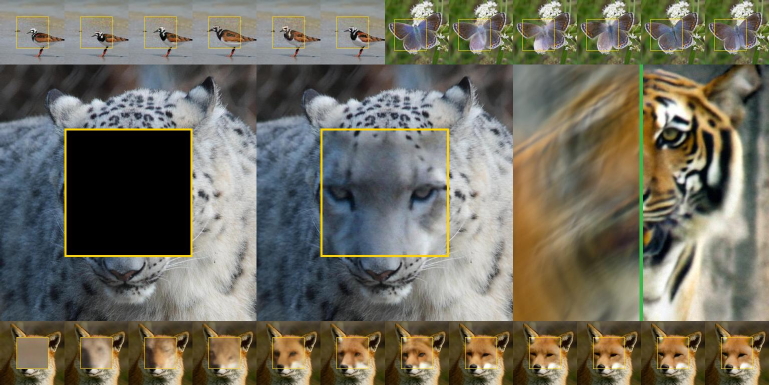}\par
  \small (b)
\end{minipage}\par
\vspace{-6pt}
\caption{
\textbf{Qualitative results.}
\textbf{(a)} $4\times$ super-resolution comparisons
with LatentDAPS and ReSample illustrate richer textures and
lower LPIPS from sampling versus higher PSNR/SSIM from Mode.
\textbf{(b)} Reconstructions on AFHQ $512\times512$ with box inpainting: the top row shows generation diversity on ImageNet samples and the bottom row shows the convergence trajectory.
}
\label{fig:qualitative}
\vspace{-12pt}
\end{figure*}

\section{Method}
\label{sec:method}
Figure~\ref{fig:Cover} summarizes CLIMB-Flow. CLIMB-Flow encodes the prior through the pretrained, stage-wise velocity model $v_\theta(x_\tau^k,\tau,k)$, which observes the noisy interpolant $x_\tau^k$ rather than the clean image $x_1^k$ itself. To sample $x_1^k$ from the stage-wise posterior $p(x_1^k\mid y)\propto p(y\mid x_1^k)p_1^k(x_1^k)$, we instead sample from a coupled target $\pi(x_1^k,x_\tau^k\mid y)$ whose $x_1^k$-marginal recovers this posterior. We use Gibbs sampling, which alternates between the coupled target's two conditionals under two Gaussian assumptions. The interpolant relation~\eqref{eq:interpolant} makes the interpolant conditional exactly Gaussian, while we approximate the
denoising factor $\pi(x_1^k\mid x_\tau^k)$ by a Gaussian with $\hat x_1^k = E[x_1^k\mid x_\tau^k]$:
\begin{eqnarray}
\pi(x_\tau^k\mid x_1^k,y)&=&\mathcal N(H_\tau^k x_1^k,(\sigma_\tau^k)^2I)\\
\label{eq:gauss_surrogate}
\pi(x_1^k\mid x_\tau^k)&\approx& \mathcal N(x_1^k;\mathbb E[x_1^k\mid x_\tau^k],C_\tau^k).
\end{eqnarray}
Here, $(C_\tau^k)^{-1}=(\sigma_\tau^k)^{-2}(H_\tau^k)^\top H_\tau^k+S_k^{-1}$ combining the interpolant precision with a clean-image covariance surrogate $S_k\succ0$ (Sec.~\ref{sec:sampling_updates}), and combine it with the measurement likelihood to obtain $\widetilde\pi(x_1^k\mid x_\tau^k,y)$ (Sec.~\ref{sec:sampling_updates}).

\begin{algorithm}[t]
\caption{CLIMB-Flow}
\label{alg:climbflow}
\small
\begin{algorithmic}[1]
\Require $y,A,\eta,v_\theta$
\For{$k=0,\ldots,K-1$}
    \State $A_k\gets A\circ U^{K-1-k}$;
    draw $x_0^k\sim\mathcal{N}(0,I_{n_k})$
    \For{$\tau\in\mathcal{T}_k$ with $\sigma_\tau^k>0$}
        \State Form $C_\tau^k,M_\tau^k$
        by~\eqref{eq:image_precision}; set $x_\tau^k$ by~\eqref{eq:interpolant}
        \For{$r=1,\ldots,S_{\mathrm{it}}$}
            \State Solve~\eqref{eq:noise_reference_estimate} for
            $\hat{x}_1^k$ using $v_\theta(x_\tau^k,\tau,k)$
            \State Solve $M_\tau^kx_1^k=b_\tau^k+\zeta$ for
            $x_1^k$ by CG
            \State Draw $x_0^k\sim\mathcal{N}(0,I_{n_k})$; re-noise
            $x_\tau^k$ by~\eqref{eq:interpolant}
        \EndFor
    \EndFor
    \State \textbf{if} $k<K-1$: $x_1^{k+1}\gets U^1(x_1^k)$
\EndFor
\State \Return $x_1^{K-1}$
\end{algorithmic}
\end{algorithm}

\vspace{-6pt}
\subsection{Clean-Endpoint Estimation}
\vspace{-3pt}
\label{sec:endpoint_estimation}
\noindent We recover $\widehat x_1^k$, the mean of the Gaussian denoising surrogate in \eqref{eq:gauss_surrogate}, from the network output $v(x_\tau^k)$ in two steps. The interpolant~\eqref{eq:interpolant} and displacement~\eqref{eq:displacement} are two linear equations in $x_1^k,x_0^k$; since the network approximates $d_\tau^k$, not $x_1^k$, and only $x_0^k$ has a known prior, $\mathcal{N}(0,I)$, we first eliminate $x_1^k$ between them, turning the network's output into a denoisable observation of $x_0^k$, then convert the result back to $x_1^k$. Throughout, we use that $H_\tau^k$, $B_k$, and $N_k:=\Delta_kH_\tau^k+\sigma_\tau^kB_k$ are all polynomials in $G$ and hence mutually commute (Sec.~\ref{sec:background}).

\noindent\textbf{Estimating $x_0^k$.} 
Scaling~\eqref{eq:interpolant} by $B_k$ gives $B_kx_\tau^k=B_kH_\tau^kx_1^k+\sigma_\tau^kB_kx_0^k$, and scaling~\eqref{eq:displacement} by $H_\tau^k$ gives $H_\tau^kd_\tau^k=H_\tau^kB_kx_1^k-\Delta_kH_\tau^kx_0^k$; since $H_\tau^k$ and $B_k$ commute, subtracting cancels $x_1^k$,
\vspace{-3pt}
\begin{equation}
B_kx_\tau^k-H_\tau^kd_\tau^k=(\sigma_\tau^kB_k+\Delta_kH_\tau^k)x_0^k=N_kx_0^k.
\vspace{-6pt}
\end{equation}
The network does not return $d_\tau^k$ exactly,so substituting its noisy estimate $v(x_\tau^k)=d_\tau^k+\varepsilon_v$ ($\varepsilon_v\sim\mathcal{N}(0,\gamma^2I)$ independent of $X_0^k$, $\gamma^2=\gamma^2(k,\tau)$ estimated from held-out pairs) gives a linear Gaussian observation of the noise endpoint,
\begin{equation}
r_\tau^k:=B_kx_\tau^k-H_\tau^kv(x_\tau^k)=N_kx_0^k-H_\tau^k\varepsilon_v,
\end{equation}
where a direct calculation gives the $\tau$-independent closed form $N_k=e_k(1-s_k)I-s_k(1-e_k)G$, computed once per stage. Combining the prior $x_0^k\sim\mathcal{N}(0,I)$ with this observation, $\widehat x_0^k$ minimizes the weighted least-squares objective
\begin{equation}
\|x_0^k\|^2+\gamma^{-2}\big\|(H_\tau^k)^{-1}(r_\tau^k-N_kx_0^k)\big\|^2,
\end{equation}
giving, after setting the gradient to zero,
$\widehat x_0^k=N_k[N_k^2+\gamma^2(H_\tau^k)^2]^{-1}r_\tau^k$.

\noindent\textbf{Back-substitution.} For $\tau>0$, $H_\tau^k$ is invertible (Sec.~\ref{sec:background}), so~\eqref{eq:interpolant} gives $H_\tau^k\widehat x_1^k=x_\tau^k-\sigma_\tau^k\widehat x_0^k$. Substituting $\widehat x_0^k$ yields
\vspace{-6pt}
\begin{equation}
\left[N_k^2+\gamma^2(H_\tau^k)^2\right]\widehat x_1^k
=\left(\Delta_kN_k+\gamma^2H_\tau^k\right)x_\tau^k+\sigma_\tau^kN_kv_\theta(x_\tau^k).
\label{eq:noise_reference_estimate}
\end{equation}
Since $0\preceq G\preceq I$ implies $N_k\succeq\Delta_kI$, the system matrix is bounded below by $\Delta_k^2I$ and remains positive definite, including at $\tau=0$. We compute $\widehat x_1^k$ using conjugate gradients.
\vspace{-6pt}
\subsection{Coupled Gaussian Updates}
\vspace{-3pt}
\label{sec:sampling_updates}
Both factors of the image conditional are now Gaussian: the likelihood $p(y\mid x_1^k)$ by the measurement model of
Sec.~\ref{sec:background}, and the denoising conditional by the surrogate $\mathcal N(\mathbb E[x_1^k\mid x_\tau^k],C_\tau^k)$ (Sec.~\ref{sec:method}), whose mean $\hat{x}_1^k$ is supplied
by~\eqref{eq:noise_reference_estimate}. Their product is Gaussian so $x_1^k$ can be drawn exactly once the surrogate covariance is fixed; $C_\tau^k$ remains positive definite even when
$H_\tau^k$ is singular. To control stochastic variability in reconstruction, we introduce
$C_{\tau,\lambda}^k=C_\tau^k/\lambda$, $\lambda\geq1$, and take
\begin{equation}
\widetilde\pi(x_1^k\mid x_\tau^k,y)\approx\mathcal{N}\big((M_\tau^k)^{-1}b_\tau^k,\,(M_\tau^k)^{-1}\big),
\label{eq:image_precision}
\end{equation}
with $M_\tau^k:=\eta^{-2}A_k^\top A_k+(C_{\tau,\lambda}^k)^{-1}$ and $b_\tau^k:=\eta^{-2}A_k^\top y
+(C_{\tau,\lambda}^k)^{-1}\hat{x}_1^k$, $\hat x_1^k$ fixed at the current interpolant.
Since $M_\tau^k\succeq S_k^{-1}\succ0$, it stays well
defined for rank-deficient $A_k$. The distribution is sampled by solving $M_\tau^kx_1^k=b_\tau^k+\zeta$ for a Gaussian perturbation $\zeta$ with $\operatorname{Cov}(\zeta)=M_\tau^k$, built from independent standard Gaussian vectors scaled by $\eta^{-1}A_k^\top$, $\sqrt{\lambda}(\sigma_\tau^k)^{-1}(H_\tau^k)^\top$, and $\sqrt{\lambda}S_k^{-1/2}$.

Omitting this perturbation returns the conditional mean
$(M_\tau^k)^{-1}b_\tau^k$; we call this deterministic variant
\emph{mode estimation}, denoted as CLIMB-Flow (Mode) and report it alongside the sampler.
We use matrix-free conjugate gradients; $S_k^{-1}$ and
$S_k^{-1/2}$ are applied by FFT-based spectral weighting.

The interpolant conditional is then sampled exactly, by drawing
$x_0^k\sim\mathcal{N}(0,I_{n_k})$ independently of the image-update
perturbations and rebuilding $x_\tau^k$
from~\eqref{eq:interpolant}. The sampled pair $(x_1^k,x_0^k)$ is
carried into the next sweep and on to the next $\tau$.

For the covariance surrogate, we take $S_k$ stationary and diagonal in the DFT: from clean reference images $\{x_{1,j}^k\}_{j=1}^J$ separate from the test set, with $\mu_k=J^{-1}\sum_j x_{1,j}^k$ and $F_k$ the unitary DFT, we set $S_k=F_k^*\operatorname{diag}(P_k)F_k$ with periodogram $P_k(\omega)=J^{-1}\sum_j|[F_k(x_{1,j}^k-\mu_k)]_\omega|^2>0$, computed once per scale, so $S_k^{-1}$ and $S_k^{-1/2}$ apply as pointwise multiplications in the transform domain.

\section{Experiments and Results}
\label{sec:exp}
\noindent\textbf{Implementation:} We build on the pretrained PixelFlow
model~\cite{chen2025pixelflow}, which uses a DiT
architecture~\cite{peebles2023scalable} and is trained on
ImageNet~\cite{deng2009imagenet}. We use
4 stages with 10 inference steps each, guidance scale
\(2.0\), and measurement noise \(\eta=0.05\) as default configuration. The sampler has two hyperparameters, the surrogate temperature $\lambda = 2$
(Sec.~\ref{sec:sampling_updates}) and
the number of Gibbs sweeps $S_{\mathrm{it}} = 2$, which trades cost for
refinement at early stages. To test pixel-domain scalability we pretrain three further PixelFlow models: AFHQ at \(512\times512\),  fastMRI~\cite{zbontar2018fastmri} at \(384\times384\), and CelebA at \(128\times128\). Each training took at most 24 GPU hours on 4 NVIDIA H100s, and inference runs on 4 NVIDIA A100s.

\noindent\textbf{Protocol:} We evaluate random and box inpainting, Gaussian deblurring and bicubic super-resolution on ImageNet and CelebA, and MRI reconstruction at $8\times$ acceleration. We report PSNR and SSIM for distortion, and LPIPS and FID for perceptual quality. Measurements and baseline numbers are taken from DAPS~\cite{zhang2025improving} and FLOWER~\cite{pourya2025flower} (random inpainting: $\eta = 0.01$), so the measurement tasks in Table~\ref{tab:combined_results} are made under same settings.
\vspace{-6pt}
\subsection{Results}

\noindent\textbf{ImageNet {256x256}:} Against pixel- and latent-domain diffusion
baselines (Table~\ref{tab:combined_results}, top left), CLIMB-Flow
attains the highest SSIM and at least the second highest PSNR on four of the five tasks, 
the lowest LPIPS and FID on box inpainting, random inpainting and super-resolution. Due to page constraints, we omit the ImageNet Gaussian and motion deblurring results from Table~\ref{tab:combined_results}. For Gaussian deblurring, Ours (Mode) achieves the second-best PSNR (26.60 dB) and SSIM (0.695). For motion deblurring, it achieves the highest SSIM (0.792) and second-best PSNR (27.57 dB), while Ours attains the second-best LPIPS (0.207) and FID (63.41).

\noindent\textbf{CelebA 128x128:} CLIMB-Flow attains the highest PSNR and SSIM on
all four tasks, improving PSNR by $1.37$--$7.66$\,dB over the competing methods
(Table~\ref{tab:combined_results}, top right).

\noindent\textbf{Qualitative comparison.} On the $4\times$ super-resolution examples (Junco and Shetland Sheepdog, Fig.~\ref{fig:qualitative}(a)), CLIMB-Flow attains higher PSNR and SSIM and lower LPIPS than LatentDAPS and ReSample. The highlighted regions follow the reference structures more closely, with fewer spurious or fragmented textures than the latent-space reconstructions. Across these examples the sampler yields higher perceptual quality and mode estimation better structural consistency: individual draws carry richer texture and the largest sample-to-sample variation, mode estimates are smoother but
score highest on PSNR and SSIM. Fig.~\ref{fig:qualitative}(b) shows reconstruction on \textbf{AFHQ 512x512}, together with the sample diversity on box inpainting on ImageNet in the top row.
\begin{figure}[!htbp]
\vspace{-6pt}
  \centering
  \includegraphics[width=\columnwidth]{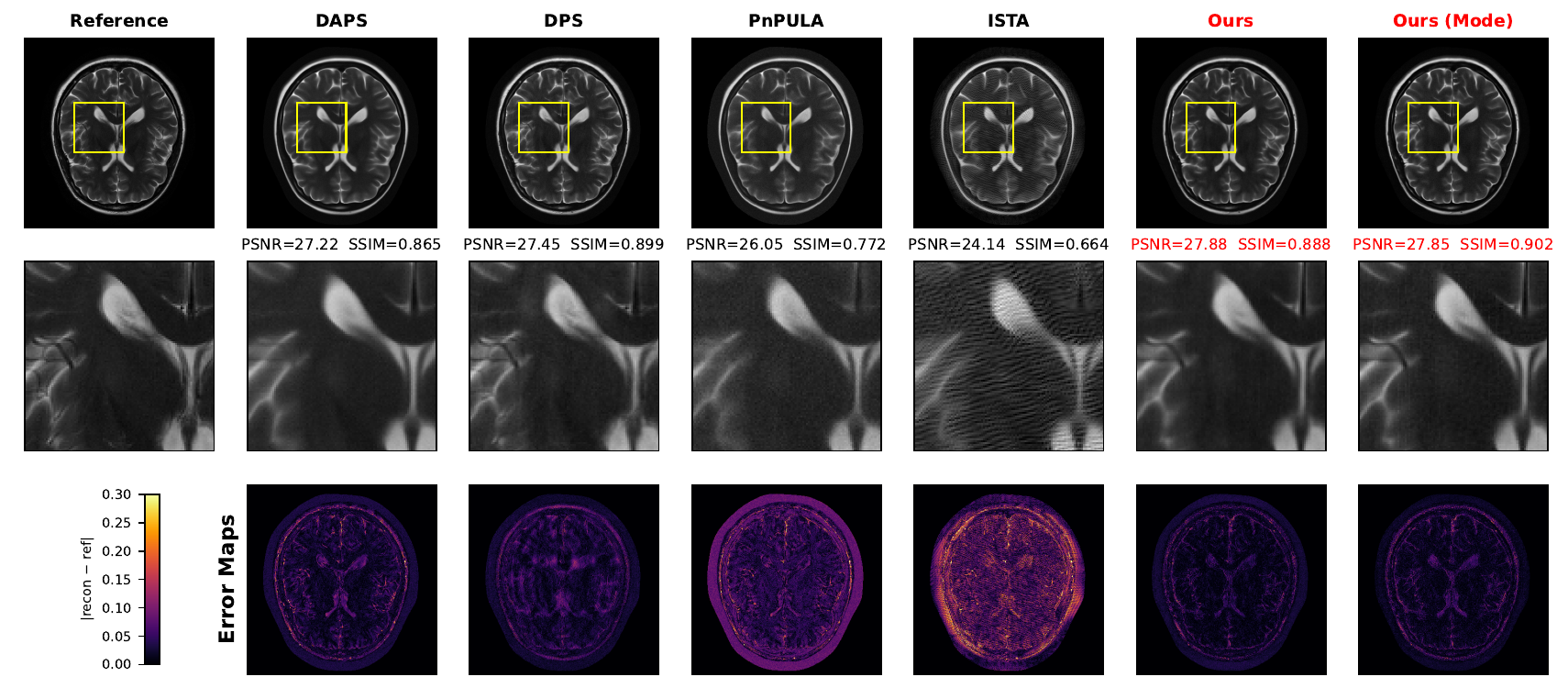}
  \vspace{-21pt}
  \caption{\textbf{fastMRI reconstruction.} Comparing with 4 baselines under $8\times$ accelerated MRI reconstruction. The MMSE and reconstruction error maps take 16 reconstructions. CLIMB-Flow algorithms achieves the best PSNR and SSIM, while preserving anatomical structure and fine-grained details.}
    \label{fig:mri}
\vspace{-6pt}
\end{figure}

\noindent\textbf{Accelerated MRI 384x384.} On the illustrated brain slice in Fig.~\ref{fig:mri} at nominal $8\times$ acceleration (53 of 384 lines), the two CLIMB-Flow variants attain PSNRs of $27.88$ and $27.85$,dB and SSIMs of $0.888$ and $0.902$, outperforming the compared baselines on both metrics. The magnified region shows better preservation of local boundaries, whereas DAPS and PnP-ULA oversmooth and ISTA introduces pronounced oscillatory artifacts. Absolute-error maps on a shared colour scale corroborate these observations, with the largest improvements over PnP-ULA and ISTA. We further evaluate single-slice MRI reconstruction on 40 test slices at nominal $8\times$ acceleration. Ours (Mode) achieves the highest mean PSNR and SSIM, reaching $24.58\pm1.56$,dB and $0.859\pm0.020$, respectively, exceeding the strongest competing baseline for each metric by $0.57$,dB and $0.017$. The stochastic variant achieves $24.40\pm1.49$,dB and $0.819\pm0.024$, ranking second in PSNR.

\noindent\textbf{Computational Analysis.}
Within four stages, CLIMB-Flow requires 80 NFE per reconstruction, of which only 20 operate at full resolution.
For the dyadic resolution pyramid, the aggregate input-pixel
count is $33.2\%$ of an equally long full-resolution schedule, while inference requires no backpropagation through the velocity network; this accounting excludes linear-solver costs.
\vspace{-6pt}
\section{Conclusions}
\vspace{-3pt}
We introduced CLIMB-Flow, a multiscale approximate flow posterior-inference framework combining shared-noise endpoint recovery with spectral Gaussian updates. It reduces repeated full-resolution network evaluations and avoids network backpropagation, while supporting reconstruction up to $512\times$. 
Its sampling and mode variants explores structural and perceptual quality across the  restoration tasks.
\vspace{-6pt}
\section{ACKNOWLEDGMENT}
\vspace{-3pt}
This work is supported by NIH grants R01-AG067078, R01-
EB031169, and R01-EB019961. The funding organization
had no role in the design, conduct, analysis, or publication of
this research.

\vfill\pagebreak




\bibliographystyle{IEEEbib_short}
\bibliography{strings,refs}

\end{document}